\documentclass{article}

\usepackage{microtype}
\usepackage{graphicx}
\usepackage{subfigure}
\usepackage{booktabs} 
\usepackage{breqn}
\usepackage{comment}

\usepackage[colorlinks=true, linkcolor=black, citecolor=blue, urlcolor=purple]{hyperref}

\usepackage{etoolbox}

\makeatletter
\patchcmd{\@footnotemark}{\Hy@raisedlink{\hyper@anchorstart{footnote.\@thefnmark}\hyper@anchorend}}{}{}{}
\patchcmd{\@makefntext}{\Hy@raisedlink{\hyper@anchorstart{footnote.#1}\hyper@anchorend}}{}{}{}
\makeatother

\PassOptionsToPackage{numbers, compress}{natbib}

\usepackage[final]{neurips_2025}

\usepackage{amssymb}
\usepackage{amsmath}
\usepackage{amsthm}
\usepackage{multirow}

\usepackage[dvipsnames]{xcolor}

\usepackage{algorithm}
\usepackage{pseudo}

\title{CoSupFormer : A Contrastive Supervised learning approach for EEG signal Classification}

\author{%
  Davy Darankoum \\
  Univ. Grenoble Alpes, CNRS, Grenoble INP, LJK, \\38000 Grenoble, France\\
  SynapCell SAS, ZAC ISIPARC, 38330 Saint-Ismier, France\\
  \texttt{davy.darankoum@univ-grenoble-alpes.fr} \\
  \And
  Chloé Habermacher \\
  SynapCell SAS, ZAC ISIPARC \\
  38330 Saint-Ismier, France \\
  \texttt{chabermacher@synapcell.fr}\\
  \And
  Julien  Volle \thanks{These authors jointly supervised the work.}\\
  SynapCell SAS, ZAC ISIPARC \\
  38330 Saint-Ismier, France \\
  \texttt{jvolle@synapcell.fr} \\
  \And
  Sergei Grudinin $^*$\\
  Univ. Grenoble Alpes, CNRS, Grenoble INP, LJK \\
   38000 Grenoble, France \\
  \texttt{sergei.grudinin@univ-grenoble-alpes.fr} \\
}

\begin{document}

\maketitle








\begin{abstract}
Electroencephalography signals (EEGs) contain rich multi-scale information crucial for understanding brain states, with potential applications in diagnosing and advancing the drug development landscape. However, extracting meaningful features from raw EEG signals while handling noise and channel variability remains a major challenge. This work proposes a novel end-to-end deep-learning framework that addresses these issues through several key innovations. First, we designed an encoder capable of explicitly capturing multi-scale frequency oscillations covering a wide range of features for different EEG-related tasks. Secondly, to model complex dependencies and handle the high temporal resolution of EEGs, we introduced an attention-based encoder that simultaneously learns interactions across EEG channels and within localized {\em patches} of individual channels. We integrated a dedicated gating network on top of the attention encoder to dynamically filter out noisy and non-informative channels, enhancing the reliability of EEG data. The entire encoding process is guided by a novel loss function, which leverages supervised and contrastive learning, significantly improving model generalization. We validated our approach in multiple applications, ranging from the classification of effects across multiple Central Nervous System (CNS) disorders treatments to the diagnosis of Parkinson's and Alzheimer's disease. Our results demonstrate that the proposed learning paradigm can extract biologically meaningful patterns from raw EEG signals across different species, autonomously select high-quality channels, and achieve robust generalization through innovative architectural and loss design.
\end{abstract}


\section{Introduction}





\paragraph{EEG signals} Electroencephalography (EEG) records the electrical activity produced by the ionic currents resulting from simultaneous activation of multiple neurons in the brain \cite{baskaran2012neurobiology,gil2023discover}. Modern medical and pharmaceutical research increasingly relies on EEG data to probe brain activity across a variety of conditions, from diagnosing neurological disorders to studying the effects of pharmacological treatments \cite{he2018brain}. The temporal and spectral complexity of EEG signals makes them a valuable source of information but also poses challenges for interpretation, particularly when analyzing raw recordings \cite{ellis2022systematic}. As the use of EEG expands across different species and research, the demand for flexible and automated feature extraction methods becomes increasingly urgent.
The richness and cross-domain relevance underscores the potential of EEG data to address a wide array of scientific and clinical questions. However, realizing this potential is challenging without effective methods for extracting informative features from raw EEG data. Traditional hand-crafted feature extraction techniques often depend on domain expertise \cite{al2018epileptic}, are specific to particular tasks, and struggle to generalize across different applications.

To fully unlock the richness of EEG data, we propose an efficient and flexible automatic encoding technique. It is based on a set of dilated convolutions and allows an autonomous extraction of a wide range of relevant features directly from raw EEG signals without requiring prior assumptions or manual feature engineering. We demonstrate its generalization across species and experimental settings.

\paragraph{Multiple Electrodes}
EEG signals are commonly recorded using multiple electrodes placed on the scalp or in-depth to capture brain activity from different regions. The coordination or disruption of activity across specific channels (each corresponding to an individual electrode) can indicate the presence of a pathology or provide insight into the effects of a drug \cite{danthine2024electroencephalogram}.  
Due to the high temporal resolution of EEG, the oscillations observed within a particular channel can track the effectiveness of a treatment or monitor unusual patterns in subjects (humans or animals) brain activity. To either capture impairments across different channels or extract some patterns within a single channel, EEG signals, usually recorded through minutes, are segmented into {\em patches}. Then, different types of architecture \cite{gu2023mamba,vaswani2017attention} are used to learn interactions between these {\em patches}. Ignoring these interactions and local dependencies can lead to incomplete or suboptimal representations of the underlying neural processes.
EEG signals are particularly vulnerable to noise and artifacts from various sources, such as muscle activity, eye movements, and environmental interference. This vulnerability makes them prone to external interferences and significantly affects the signal-to-noise ratio (SNR), challenging data integrity and reliability \cite{minguillon2017trends, jas2017autoreject, bigdely2015prep}. As a result, some channels may contribute significantly to the noise while providing little or no valuable information about the underlying neural activity.
In recent EEG analysis pipelines \cite{wangmedformer, ingolfsson2020eeg}, all channels are typically treated as equally important. This assumes a consistent level of signal quality across the entire electrode array. However, this assumption frequently does not hold in real-world scenarios, especially in cases involving invasive recordings. The presence of noisy or corrupted channels makes learning robust representations from raw EEG data particularly difficult. Such channels can obscure meaningful patterns, lower the signal-to-noise ratio at the representation level, and hinder the model's ability to generalize.
A major challenge in EEG representation learning, therefore, is the identification and selective suppression or mitigation of highly noisy, redundant, or non-informative channels. 
This paper presents a novel adaptive approach that uses a combination of a gating mechanism with global attention to selectively activate only meaningful features from clean {\em patches} of multi-channel EEG data.

\section{Related Works}

\paragraph{Feature extraction from EEG signals}  
EEG signals can be manually processed to extract features in the time, frequency, and time–frequency domains, as well as through non-linear analyses \cite{durongbhan2019dementia, singh2023trends, guo2010automatic, wang2018hardware, mursalin2017automated}. While these handcrafted features, when used with traditional machine learning classifiers, have demonstrated good performance in disease diagnosis, they often fail to generalize across subjects \cite{einizade2020deep}. 
Additionally, the diversity of potential features introduces a need for task-specific selection, which can limit generalizability.
To overcome these limitations, recent research has focused on automatic feature extraction. 
Convolutional neural networks (CNNs) and recurrent neural networks (RNNs) have been employed to learn expressive representations directly from raw EEG signals or from time–frequency representations such as the short-time Fourier transform (STFT) coefficients \cite{acharya2018deep, abdulwahhab2024detection, roy2018deep, cho2020comparison}. 
For instance, \citet{eldele2021attention} proposed a multi-resolution CNN with two parallel branches using different kernel sizes to capture both high- and low-frequency components for sleep stage classification. 
Similarly,  EEGNet  developed compact depthwise and separable convolutions to classify EEG signals in different Brain Computer Interface (BCI) tasks \cite{lawhern2018eegnet}. 
Other methods segment EEG signals into multiple {\em patches} of different sizes prior to processing \cite{nietime, zhang2024multi}. These {\em patches} are then encoded using MLP or CNN-based architectures to extract multigranularity features  \cite{chenpathformer, shabani2022scaleformer, wangmedformer}.

\paragraph{EEG transformers}
Transformer-based architectures \cite{vaswani2017attention} have been widely adapted for EEG analysis due to their ability to model long-range dependencies. In EEG-based applications, attention mechanisms are typically used to learn either temporal relationships between {\em patches} or interactions between extracted features.
Most existing work has focused on learning patch-level interactions. 
We identify three types of these:
(1) interactions within {\em patches} from a single channel (within-channel interactions); (2) interactions between {\em patches} across channels at the same time step (cross-channel interactions at the same time); and (3) interactions across channels and across time (cross-channel interactions at different times).

PatchTST \cite{nietime}, MTST \cite{zhang2024multi}, and CrossFormer \cite{zhang2023crossformer} focus on within-channel interactions. While CrossFormer uses fixed-size {\em patches}, PatchTST and MTST apply multi-granularity patching, resulting in variable-sized {\em patches}. MedFormer  \cite{wangmedformer}, on the other hand, combines all channels before patching and models cross-channel interactions at a fixed time. This approach enables learning of spatiotemporal features from all channels simultaneously, distinguishing it from methods that apply patching on individual channels. EEG Conformer \cite{song2022eeg} learns both within-channel and cross-channel interactions at the same time step through a combined convolution and transformer architecture.
To the best of our knowledge, no method explicitly learns interactions between {\em patches} from different channels at different time points, despite evidence suggesting that disruptions in inter-regional brain dynamics are associated with neurological disorders \cite{guillon2017loss, elkholy2023disruption}.
Our method addresses this gap by modeling cross-channel interactions across multiple time scales in an attention-based mechanism. Within-channel interactions and cross-channel interaction at the same time step are also leveraged in a global attention mechanism. 

Some models also focus on feature-level interactions. iTransformer \cite{liuitransformer} computes attention across features from all channels, while CrossFormer applies attention within features from individual {\em patches}. More recent BioMamba  model \cite{qian2025biomamba} replaces traditional transformers with a Mamba-based architecture  \cite{gu2023mamba} to reduce memory consumption during patch interaction modeling.

\paragraph{Learning from multi-channel noisy data}
Various strategies have been proposed to mitigate the impact of noise in EEG recordings. Data augmentation techniques, tailored to EEG, have been shown to improve model generalization \cite{rommel2022data}. Common augmentation methods include time-domain jittering, frequency perturbations, and spatial channel shuffling. \citet{wang2025time} introduced a method based on discrete wavelet transform to account for EEG non-stationarity. However, most augmentation methods have not been rigorously evaluated under high noise scenarios. They also increase training resource requirements.

Graph-based architectures offer an alternative by modeling spatial relationships between electrodes to enable denoising. For example, \citet{pentari2022graph} and \citet{yang2024you} proposed graph-based denoising techniques that learn robust EEG representations. Preprocessing methods such as independent component analysis (ICA), variational mode decomposition (VMD), and wavelet-based filtering are also commonly used to suppress noise \cite{salem2025strategies, kashyap2024enhanced}.

\paragraph{Self-supervised learning}
Self-supervised learning (SSL) has emerged as a promising approach for EEG classification, enabling models to learn from large volumes of unlabeled data. EEGPT \cite{wang2024eegpt} employs a transformer-based architecture trained with dual objectives: mask-based reconstruction and spatiotemporal alignment. TS-TCC \cite{eldele2021time} introduces temporal and contextual contrastive learning to model signal dynamics, while ContraWR \cite{yang2021self} applies global contrastive learning for automatic sleep staging. These methods have demonstrated improved robustness and generalization, particularly in low-label or cross-subject scenarios. Recently, the EEG decoding field witnessed the development of foundation models like LaBraM \cite{jianglarge}, NeuroML \cite{jiangneurolm} and CBraMod \cite{wangcbramod} which all leverage the increasing number of unlabeled EEG datasets to learn rich EEG-based embeddings and improve accuracy in downstream tasks.

\paragraph{Key contributions}
\begin{itemize}
    \item We propose to combine a dilated CNN (as in \citet{yu2015multi}) with a multi-resolution CNN (as in \citet{eldele2021attention}) for feature extraction from EEG signals. Compared to \citet{eldele2021attention}, our encoder reduces the number of learnable parameters through the selective use of small kernels for high-frequency patterns and dilated large kernels for low-frequency patterns. It also demonstrates transferability across datasets and species.
    
    \item We introduce a global attention mechanism that models both within-channel and cross-channel patch interactions. Unlike MedFormer \cite{wangmedformer} and EEG Conformer \cite{song2022eeg}, our design allows interactions between {\em patches} from different channels at different time steps. Furthermore, attention weights are jointly optimized for both interaction types, improving generalization.
    
    \item To filter out noisy {\em patches} and channels, we propose a gating mechanism that works in tandem with the global attention module.
    
    \item Finally, we present a hybrid loss function that integrates supervised and contrastive objectives to leverage self-supervised learning with a reduced number of parameters compared to current foundation models.
\end{itemize}

\section{Methods}
\subsection{Multi-scale EEG Encoding}
Brain activity recorded through EEG is typically sampled at around 500 Hz, allowing for the extraction of frequency components between 0.1 and 250 Hz. This wide frequency range is critical, as it encompasses neural rhythms from low-frequency bands (e.g., 0.5–4 Hz, associated with sleep stages) to high-frequency bands (e.g., 30-100 Hz, linked to cognitive activity). These spectral components are often indicative of specific brain states or dysfunctions. Hence, developing an encoder that can automatically extract informative frequency-specific features from raw EEG is essential.

Inspired by the work of \citet{eldele2021attention}, 
we construct a dual-branch convolutional encoder. The first branch uses small convolutional kernels to focus on fine-grained, high-frequency features. The second branch uses larger kernels with dilation to capture low-frequency oscillations over broader time windows.
Unlike traditional large-kernel convolutions that tend to capture overlapping or redundant frequency information due to closely spaced receptive fields, our design employs dilated convolutions to address this issue. By inserting controlled gaps (dilations) between kernel elements, dilated convolutions expand the receptive field without increasing the number of parameters. This spacing reduces the overlap between adjacent receptive fields across layers, thereby encouraging the network to focus on distinct, non-redundant temporal patterns corresponding to specific frequency bands. As a result, our architecture promotes better frequency disentanglement and reduces model complexity by minimizing parameter count (see Figure \ref{fig:arch}).
\paragraph{Dual-path feature extraction encoder for multichannel EEGs}
Let $x \in \mathbb{R}^{C \times L}$ be a multichannel EEG signal, where $C$ denotes the number of electrodes (channels), and $L$ denotes the temporal length of the signal. 
We independently pass each channel $x_c \in \mathbb{R}^{1 \times L}$ through two distinct 1D convolutional encoders, {\em Encoder 1} and {\em Encoder 2}, to extract complementary representations.
We define two functions,
\begin{itemize}
    \item (Encoder 1) $f_1: \mathbb{R}^{1 \times L} \rightarrow \mathbb{R}^{E_1 \times L_1}$, an encoder composed of small-kernel convolutions aimed at capturing fine-grained temporal patterns and localized frequency features.
    \item (Encoder 2) $f_2: \mathbb{R}^{1 \times L} \rightarrow \mathbb{R}^{E_2 \times L_2}$, an encoder composed of large-kernel convolutions with dilations to capture broader temporal dependencies and coarse-scale features.
\end{itemize}
Each EEG channel $X_c$ is processed individually by both encoders using independent weights.
The initial feature maps $z_{1}^{(c)}$ and $z_{2}^{(c)}$ are thus,
\begin{equation}
z_{1}^{(c)} = f_1(x_c)\in \mathbb{R}^{E_1 \times L_1}, \quad z_{2}^{(c)} = f_2(x_c) \in \mathbb{R}^{E_2 \times L_2},
\end{equation}
where we fixed the size of both encoder embeddings to $E=E_1=E_2$.
Then, we concatenate the two feature maps along the sequence dimension,
\begin{equation}
    z^{(c)} = \text{Concat}_l(z_{1}^{(c)}, z_{2}^{(c)}) \in \mathbb{R}^{E \times L},
\end{equation}
where $L = L_1 + L_2$. A dropout layer is applied on the feature dimension to increase model generalization,
\begin{equation}
z^{(c)} = \text{Dropout}\left(z^{(c)}\right).
\end{equation}
The final encoded representation $z \in \mathbb{R}^{CL \times E}$ for the full EEG input is obtained by stacking the representations across all channels,
\begin{equation}
z =\left[ z^{(1)}, z^{(2)}, \dots, z^{(C)} \right].
\end{equation}

Algorithm \ref{alg:enc} schematically list one encoder branch sequence.

\begin{algorithm}
\pseudoset{
ctfont=\color{black!25},
ct-left= \hspace{0pt plus 1filll }, ct-right=,
}
\begin{pseudo*}[indent-mark]
{\color{ForestGreen} \textbf{def}} Encoder($x \in \mathbb{R}^{C \times L}$)          \\

{\color{Mahogany} {\em \# loop through channels}}    \\
for $c$ in $C$ :  \\+
    {\color{Mahogany} {\em \# input $x_c \in \mathbb{R}^{1 \times L}$}}    \\
    1: $z_{1}^{(c)} = \text{Encoder}_{1,c}(x_c)$     \hskip1.5em \ct{contains Conv1D, BatchNorm, Gelu and MaxPooling layers}\\
    2: $z_{2}^{(c)} = \text{Encoder}_{2,c}(x_c)$      \ct{contains Conv1D, BatchNorm, Gelu and MaxPooling layers}\\
    3: $z^{(c)} = \text{Concat}_l(z_{1}^{(c)}, z_{2}^{(c)})$      \ct{concatenate along the sequence dim}     \\-

{\color{Mahogany}{\em \# dropout}}    \\
4: $z^{(c)} \leftarrow \text{Dropout}\left(z^{(c)}\right)$ \\
{\color{Mahogany}{\em \# stacking outputs from all channels along the sequence dim}}    \\
5: $z = \left[ z^{(1)}, z^{(2)}, \dots, z^{(C)} \right]$ \\
6: {\color{ForestGreen} \textbf{return}} $z \in \mathbb{R}^{CL \times E}$
\end{pseudo*}
\caption{Encoder}
\label{alg:enc}
\end{algorithm}

\subsection{Gated global attention}
We propose an attention encoder which learns into a single attention matrix, within-channel interactions, cross-channel interactions at the same time, and cross-channel interactions at different time steps. Compared to previous approaches used with EEG signals \citep{wangmedformer, nietime,zhang2024multi, zhang2023crossformer}, where within-channel patch interactions and cross-channel patch interactions are learned separately, our method combines both at once. Learning interactions relevance within such a global attention mechanism allows the model to either focus on within-channel interactions and/or cross-channel interactions. Therefore, increasing the generalizable capabilities as for some tasks within-channels relationship are very important \citep{abdelhameed2021deep} and for others cross-channel interactions are more informative \citep{kundaram2020deep}.

Moreover, enabling cross-channel interactions at different time scales increases the model’s expressiveness by allowing it to capture and weight potential disparities observed through different channels. Finally, this global attention design also provides a foundation for improving the suppression of noisy {\em patches} (see details below).

\paragraph{Global Attention module}
Let $z \in \mathbb{R}^{CL \times E}$ be the output of the dual-path feature extraction encoder, where $C$ is the number of EEG channels, $L$ is the sequence length per channel and $E$ the embedding dimension. $C L$ represents the full set of spatio-temporal positions across all channels. This flattened representation serves as input to our attention module. Our network is equivariant with respect to the dimension $L$. The attention matrix, therefore depends on the initial input size of $L$. 
To model spatio-temporal interactions, we apply a self-attention mechanism with rotary positional embedding \citep{su2024roformer} (see Algorithm \ref{alg:att} for more details).
We compute the attention coefficients $a$ as follows. We remove diagonal elements to avoid self-attention and apply a row-wise softmax,
\begin{equation}
    a = \text{softmax}_i(\text{mask}(A)),
\text{with mask}(A) \equiv A - \text{diag}(A).
\end{equation}

\begin{figure}
    \centering
    \includegraphics[width=1\linewidth]{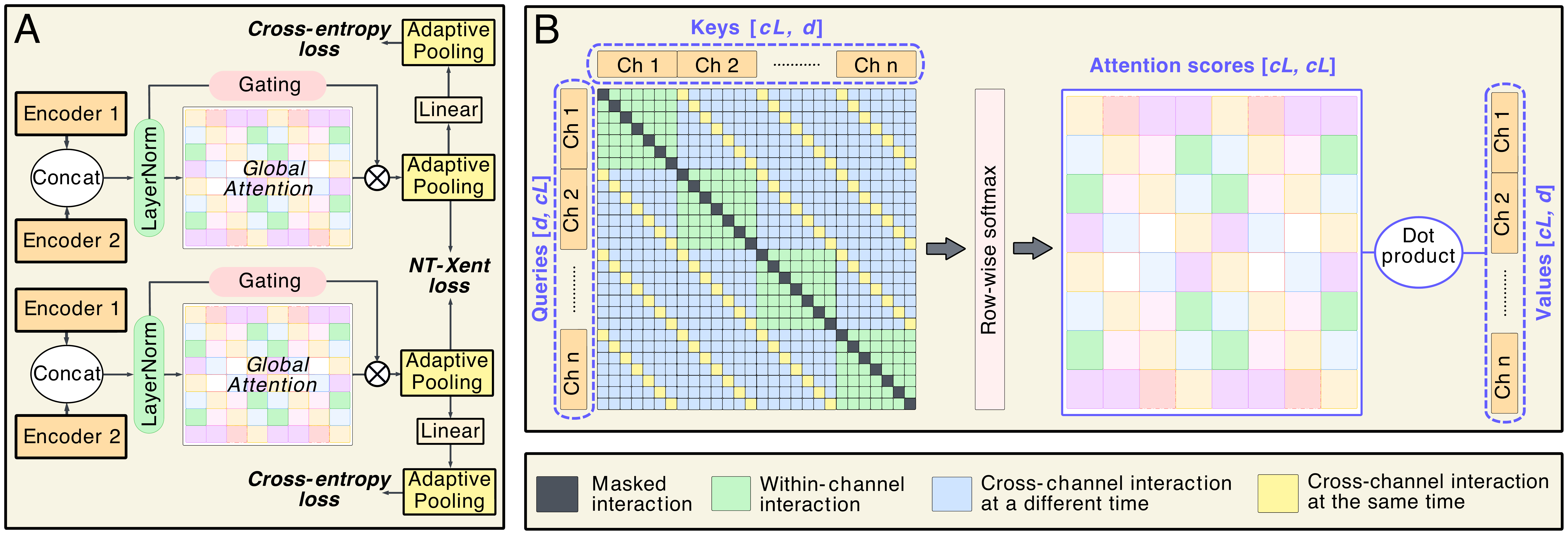}
    \caption{
    A) Schematic illustration of the CoSupFormer architecture. B) The global attention module, where $c$ denotes the number of channels, $L$ the sequence length per channel, and $d$ the embedding dimension. This module is applied multiple times, once for each attention head.
    }
    \label{fig:arch}
\end{figure}

Finally, the output is projected back and passed through residual and feed-forward layers, as outlined in Algorithm \ref{alg:att}.
See Fig. \ref{fig:arch}B for a schematic illustration of the module. 

\begin{algorithm}
\pseudoset{
ctfont=\color{black!25},
ct-left= \hspace{0pt plus 1filll}, ct-right=,
}
\begin{pseudo*}[indent-mark]
{\color{ForestGreen} \textbf{def}} GlobalGatedAttention$\left(z \in \mathbb{R}^{CL \times E}\right)$                               \\+
{\color{Mahogany} {\em \# input projections}}    \\
1: $\bar z = \text{LayerNorm} (z)$     \\
2: $q,k,v = \text{LinearNoBias}(\bar z)$         \\
3: $q \leftarrow \text{RoPE}(q)$, \quad  $k \leftarrow \text{RoPE}(k)$           \\
4: $g = \text{sigmoid}\left(\text{Linear}(\bar z)\right)$        \\

{\color{Mahogany}{\em \# row masked attention}}     \\
5: $a = \text{softmax}_i(\text{mask}({\frac{1}{\sqrt{d}}}q^Tk))$
\qquad \qquad \quad \ct{ $\text{mask}(A) \equiv A - \text{diag}(A)$ and $d = E/\#Heads$} \\
{\color{Mahogany}{\em \# update of z }}  \\
6: $z \leftarrow \text{LayerNorm} (a v + \bar z) $     \\
{\color{Mahogany}{\em \# residual feed-forward network}}     \\
7: $z \leftarrow \text{LayerNorm} \left(z + \text{FFN}(z)\right) $     \\
{\color{Mahogany}{\em \# output projections}}    \\
8: $o = \text{Linear}(z)$        \\
{\color{Mahogany}{\em \# gating}}     \\
9: $o \leftarrow o \odot g$     \\
10: {\color{ForestGreen} \textbf{return}} $o \in \mathbb{R}^{CL \times E}$

\end{pseudo*}
\caption{Global gated attention}
\label{alg:att}
\end{algorithm}

\paragraph{Gating}
While the attention mechanism already reduces the impact of non-informative {\em patches} by assigning them low interaction weights, these {\em patches} still remain in the feature space and can influence downstream decisions. 
To address this limitation, we introduce an additional gating mechanism that explicitly suppresses noisy or irrelevant patch representations.
The gating network operates as a learned element-wise mask. 
Given the output of the dual-path feature extraction encoder, we first normalize it, then project and squeeze the obtained representation through a sigmoid function to obtain gating weights, as outlined in Algorithm \ref{alg:att} (line 4). The final gated representation is computed via element-wise multiplication, see Algorithm \ref{alg:att} (line 9).
This gating mechanism allows the model to completely suppress the contribution of irrelevant or noisy {\em patches}, particularly those that the attention mechanism has already deprioritized. 
Unlike soft attention, which merely downweights poor interactions, the gating mechanism enforces a stronger, task-driven filtering. The combination of attention-based scoring and explicit gating ensures that noisy or misleading information is entirely removed from the model's final decision-making process.

\subsection{CoSup loss : Contrastive and Supervised loss}
To address the persistent challenge of poor generalization in deep learning models applied to EEG data, we propose a novel loss function that enhances the model’s ability to learn robust representations. EEG signals are known to exhibit significant variability due to temporal fluctuations, inter-subject differences, and inconsistencies in electrode placement or configuration. As a result, models trained on such data often struggle to maintain performance across sessions, subjects, or recording setups. For practical applicability in real-world scenarios, a model must therefore demonstrate invariance to temporal shifts and channel permutations.

Let $f(x_i) \in \mathbb{R}^{L \times E}$ be the learned representation of an EEG sample $x_i$, where $L$ is the sequence length and $E$ is the embedding dimension.
Assuming a supervised learning setting for a $K$-class classification problem, 
let each sample $x_i$ have a corresponding ground-truth label $y_i \in \{0, 1, \dots, K-1\}$.
We then introduce a hybrid loss function $\mathcal{L}$  that combines the standard multi-class cross-entropy loss $\mathcal{L}_{\text{CE}}$ with a contrastive learning objective $\mathcal{L}_{\text{NT-Xent}}$,
\begin{equation}
    \mathcal{L} = \lambda  \mathcal{L}_{\text{CE}} + (1 - \lambda)  \mathcal{L}_{\text{NT-Xent}}. 
\end{equation}
%
%
The combination of two objectives leads to more discriminative and generalizable feature representations. A single weighting parameter $\lambda$ balances the contributions of the two terms during training, allowing fine-grained control over the learning dynamics.

\paragraph{Contrastive loss component $\mathcal{L}_{\text{NT-Xent}}$}
We first compute the pairwise cosine similarity matrix $s_{ij}$,
\begin{equation}
    s_{ij} = \frac{z_i^\top z_j}{\|z_i\| \|z_j\|}, \quad \text{for } i, j \in \{0,1, \dots, B\},
\end{equation}
where $z_i \in \mathbb{R}^{L E}$ is the flattened embedding of $f(x_i)$, and
$B$ is the total batch size. 
We then define a binary label matrix,
\begin{equation}
M_{ij} = 
\begin{cases}
1 & \text{if } y_i = y_j \text{ and } i < j \\
0 & \text{otherwise}
\end{cases}.
\end{equation}
Let $\tau$ be a temperature scaling parameter. The NT-Xent loss is computed over the positive pairs (i.e., pairs with the same label, e.g., the upper triangular part of the binary label matrix $M_{ij}$) as:
\begin{equation}
\mathcal{L}_{\text{NT-Xent}} = - \frac{1}{|P|} \sum_{(i,j) \in P} \log \left( \frac{\exp(s_{ij}/\tau)}{\sum_{k \neq i} \exp(s_{ik}/\tau)} \right),
\end{equation}
where $P = \{(i,j) \mid M_{ij} = 1\}$ is the set of valid positive pairs in the batch \citep{chen2020simple}.
\paragraph{Cross-Entropy loss component $\mathcal{L}_{\text{CE}}$}
To compute the cross-entropy loss, we 
first average
the temporal dimension of the embeddings,
$
\bar{z}_i = \text{AvgPool}(f(x_i)) \in \mathbb{R}^E
$.
Then we apply a softmax to convert the pooled embedding into a probability distribution over classes:
\begin{equation}
p(y \mid \bar{z}_i) = \text{softmax}(\bar{z}_i).
\end{equation}
The cross-entropy loss is computed as:
\begin{equation}
\mathcal{L}_{\text{CE}} = - \frac{1}{B} \sum_{i=1}^{B} \log p(y_i \mid \bar{z}_i),
\end{equation}
where $p(y_i \mid \bar{z}_i)$ is the predicted probability for the true class $y_i$.



\section{Results and Discussion}
\label{sec:results}
We evaluated our methods on five datasets : TDBrain, TDBrain+Noise, ADFTD, ADFTD+Noise, and MACO, and compared with five baseline architectures : iTransformer  \cite{liuitransformer}, FEDFormer \cite{zhou2022fedformer}, PatchTST \cite{nietime}, CrossFormer \cite{zhang2023crossformer} and MedFormer \cite{wangmedformer}. These baselines were selected to cover a wide range of state-of-the-art transformer-based models. They include models that capture both patch-level and feature-level interactions (CrossFormer, MedFormer), as well as those focused solely on feature interactions (iTransformer). Additionally, we selected PatchTST and FEDFormer as strong representatives of transformer models specifically designed for time series forecasting.

\subsection{Datasets}
\paragraph{The TDBrain dataset} TDBrain is a large EEG time series collection introduced in \citet{van2022two}, with recorded brain activities from 1,274 subjects across 33 channels.
Each subject participated in two trials: eyes-closed and eyes-open conditions.
The dataset contains a total of 60 labels, allowing for the possibility of multiple labels per subject to indicate coexisting diseases.
Following \citet{wangmedformer}, we used a subset of TDBrain that includes 25 subjects diagnosed with Parkinson’s disease and 25 healthy controls, all under eye-closed task condition. 
%
Training, validation, and test sets have been constructed in a subject-independent manner.

\paragraph{The TDBrain+Noise dataset} 
To assess the capabilities of our gated global attention, we constructed the TDBrain+Noise dataset by {\em noising} one third of the channels from the original TDBrain data. 
More concretely, we added Gaussian noise to the last 11 of 33 channels. Details on the noising process are provided in the Supplementary Material section \ref{TDBrain_Preproc}. 

\paragraph{The ADFTD dataset} 
The Alzheimer’s Disease and Frontotemporal Dementia (ADFTD) dataset is a publicly available EEG time series dataset comprising recordings from three subject groups: 36 patients with Alzheimer’s Disease (AD), 23 with Frontotemporal Dementia (FTD), and 29 healthy control (HC) individuals \citep{miltiadous2023dataset}. Each subject underwent a single EEG recording session using 19 channels, captured at a native sampling rate of 500 Hz. The recording durations vary across groups, with AD sessions averaging approximately 13.5 minutes, FTD sessions around 12 minutes, and HC sessions about 13.8 minutes. This variability reflects natural clinical differences in session lengths and provides a rich temporal structure for analysis.
For evaluation, a subject-independent split is used, where subjects are divided into training (60\%), validation (20\%), and test (20\%) sets to ensure no subject appears in multiple subsets.

\paragraph{The ADFTD+Noise dataset} 
To evaluate the robustness of our model under degraded signal conditions, we derived the ADFTD+Noise dataset by {\em noising} a subset of the EEG channels from the original ADFTD recordings. Specifically, we injected Gaussian noise into the last 6 of the 19 available channels. Details of the noise injection process are provided in the Supplementary section \ref{ADFTD_Preproc}.

\paragraph{The MACO dataset} MACO is a private dataset comprising 1536 hours of EEG recordings from 128 mice. It includes multiple pharmaco-EEG studies involving compounds from two therapeutic classes: antidepressants and antipsychotics. A third class consists of EEG recordings obtained after solvent administration, included to constitute a control condition. Signals were recorded from three brain regions, the hippocampus, the prefrontal cortex, and the parietal cortex, resulting in 3-channel EEG data. All experiments followed a cross-over design to enhance data consistency and reduce the number of animals required. As the recordings are performed on animals, the dataset naturally includes artifacts related to movements and other biological sources of noise.
All the signals were acquired at 1024Hz and resampled at 256Hz. The dataset was segmented into training, validation and test sets in a subject independent manner, as 70\%, 10\% and 20\%, respectively.

\begin{table}[ht]
\centering
{\scriptsize
\caption{Comparison of CoSupFormer with five baselines on two clean datasets.} 
\label{tab:clean}
\begin{tabular}{llcccccc}
\toprule
\textbf{Dataset} & \textbf{Model} & \textbf{Accuracy} & \textbf{Precision} & \textbf{Recall} & \textbf{F1 Score} & \textbf{AUROC} & \textbf{AUPRC} \\
\midrule

\multirow{6}{*}{TDBrain}
     & iTransformer \cite{liuitransformer}   & 74.67 ± 1.06 & 74.71 ± 1.06 & 74.67 ± 1.06 & 74.65 ± 1.06 & 83.37 ± 1.14 & 83.73 ± 1.27 \\
     & FEDformer  \cite{zhou2022fedformer}    & 78.13 ± 1.98 & 78.52 ± 1.91 & 78.13 ± 1.98 & 78.04 ± 2.01 & 86.56 ± 1.86 & 86.48 ± 1.99 \\
     & PatchTST   \cite{nietime}    & 79.25 ± 3.79 & 79.60 ± 4.09 & 79.25 ± 3.79 & 79.20 ± 3.77 & 87.95 ± 4.96 & 86.36 ± 6.67 \\
     & CrossFormer \cite{zhang2023crossformer}   & 81.56 ± 2.19 & 81.97 ± 2.25 & 81.56 ± 2.19 & 81.50 ± 2.20 & 91.20 ± 1.78 & 91.51 ± 1.71 \\
     & MedFormer \cite{wangmedformer}      & 89.62 ± 0.81 & 89.68 ± 0.78 & 89.62 ± 0.81 & 89.62 ± 0.81 & 96.41 ± 0.35 & 96.51 ± 0.33 \\
     & CoSupFormer    & \textbf{95.81 ± 0.41} & \textbf{97.30 ± 2.28} & \textbf{94.33 ± 2.70} & \textbf{95.74 ± 0.47} & \textbf{99.45 ± 0.07} & \textbf{99.50 ± 0.06} \\

\midrule

\multirow{6}{*}{ADFTD}
     & iTransformer \cite{liuitransformer}  & 52.60 ± 1.59 & 46.79 ± 1.27 & 47.28 ± 1.29 & 46.79 ± 1.13 & 67.26 ± 1.16 & 49.53 ± 1.21 \\
     & FEDformer \cite{zhou2022fedformer}     & 46.30 ± 0.59 & 46.05 ± 0.76 & 44.22 ± 1.38 & 43.91 ± 1.37 & 62.62 ± 1.75 & 46.11 ± 1.44 \\
     & PatchTST   \cite{nietime}    & 44.37 ± 0.95 & 42.40 ± 1.13 & 42.06 ± 1.48 & 41.97 ± 1.37 & 60.08 ± 1.50 & 42.49 ± 1.79 \\
     & CrossFormer \cite{zhang2023crossformer}   & 50.45 ± 2.31 & 45.57 ± 1.63 & 45.88 ± 1.82 & 45.50 ± 1.70 & 66.45 ± 2.03 & 48.33 ± 2.05 \\
     & MedFormer \cite{wangmedformer}     & \textbf{53.27 ± 1.54} & \textbf{51.02 ± 1.57} & \textbf{50.71 ± 1.55} & \textbf{50.65 ± 1.51} & \textbf{70.93 ± 1.19} & 51.21 ± 1.32 \\
     & CoSupFormer    & \textbf{54.95 ± 1.80} & \textbf{51.18 ± 1.07} & \textbf{51.26 ± 0.87} & \textbf{50.88 ± 0.78} & \textbf{70.43 ± 0.57} & \textbf{53.72 ± 0.67} \\

\bottomrule
\end{tabular}
} 
\end{table}

\begin{table}[ht]
\centering
{\scriptsize
\caption{Comparison of CoSupFormer with two baselines on three challenging datasets.}
\label{tab:noisy}
\begin{tabular}{llcccccc}
\toprule
\textbf{Dataset} & \textbf{Model} & \textbf{Accuracy} & \textbf{Precision} & \textbf{Recall} & \textbf{F1 Score} & \textbf{AUROC} & \textbf{AUPRC} \\
\midrule

\multirow{3}{*}{TDBrain+Noise}
     & CrossFormer \cite{zhang2023crossformer}  & 68.77 ± 1.30 & 66.03 ± 3.04 & 78.29 ± 5.85 & 71.44 ± 1.24 & 77.60 ± 1.98 & 78.11 ± 1.96 \\
     & MedFormer \cite{wangmedformer}    & 49.79 ± 1.37 & 49.85 ± 0.93 & 74.29 ± 3.04 & 59.65 ± 1.61 & 50.27 ± 0.38 & 50.61 ± 1.18 \\
     & CoSupFormer   & \textbf{94.98 ± 1.94} & \textbf{93.86 ± 3.99} & \textbf{96.46 ± 1.59} & \textbf{95.09 ± 1.75} & \textbf{99.31 ± 0.18} & \textbf{99.37 ± 0.20} \\

\midrule

\multirow{3}{*}{ADFTD+Noise}
     & CrossFormer \cite{zhang2023crossformer}  & 49.88 ± 1.71 & 46.96 ± 1.17 & 45.17 ± 0.63 & 45.12 ± 0.59 & 63.66 ± 0.63 & 46.81 ± 0.57 \\
     & MedFormer  \cite{wangmedformer}   & 37.63 ± 0.34 & 33.53 ± 0.30 & 33.48 ± 0.25 & 33.15 ± 0.24 & 50.07 ± 0.19 & 33.42 ± 0.20 \\
     & CoSupFormer   & \textbf{51.20 ± 1.57} & \textbf{47.62 ± 1.95} & \textbf{46.50 ± 0.70} & \textbf{46.25 ± 0.85} & \textbf{66.73 ± 0.50} & \textbf{50.96 ± 0.69} \\

\midrule

\multirow{3}{*}{MACO}
     & CrossFormer \cite{zhang2023crossformer}  & 38.82 ± 11.18 & 12.94 ± 3.73 & 33.33 ± 0.00 & 18.41 ± 3.89 & 50.04 ± 0.00 & 33.55 ± 0.01 \\
     & MedFormer  \cite{wangmedformer}   & 32.29 ± 11.28 & 10.76 ± 3.76 & 33.33 ± 0.00 & 16.01 ± 4.07 & 49.98 ± 0.11 & 33.66 ± 0.08 \\
     & CoSupFormer & \textbf{74.93 ± 4.46} & \textbf{73.30 ± 6.56} & \textbf{71.68 ± 6.12} & \textbf{71.74 ± 6.24} & \textbf{87.66 ± 4.08} & \textbf{79.52 ± 7.76} \\

\bottomrule
\end{tabular}
} 
\end{table}

\subsection{Experiments}
\label{subsec:Experiments}
\paragraph{Experiment details} 
All models were trained for 100 epochs using the PyTorch framework and the Adam optimizer. Hyperparameters such as learning rate, weight decay, and embedding dimension were tuned based on validation F1-score. We also tested various activation functions (sigmoid, tanh, softmax) in the gating mechanism and selected the best-performing variant.
Experiments were conducted on an NVIDIA Tesla V100 GPU cluster (3 nodes, 32GB RAM each).
Training time and parameter counts varied across datasets. For instance, on TDBrain+Noise, CoSupFormer trained in 332 minutes with 1.07M parameters, compared to 15 minutes for CrossFormer (5.3M parameters) and 35 minutes for MedFormer (3.5M parameters). On MACO, CoSupFormer used only 415K parameters, significantly fewer than MedFormer (3.2M) and CrossFormer (5.2M). While CoSupFormer’s training time was 271 minutes, MedFormer and CrossFormer required 580 and 420 minutes, respectively—partly due to the use of gradient accumulation to fit their larger models within our memory constraints.
These training details highlight CoSupFormer’s efficiency in model size, while still achieving superior performance. See section \ref{Exp_setup} in Supplementary Material for more details.

\paragraph{Results} 
Table \ref{tab:clean} presents the results of our experiments on datasets free from eye movement artifacts or external noise. These include TDBrain and ADFTD. We will refer to them as to {\em clean datasets}. In contrast, Table \ref{tab:noisy} shows the results on datasets that contain  either synthetic noise or biologically induced artifacts.
 These include TDBrain+Noise,  ADFTD+Noise, and MACO. We will refer to them as to {\em challenging datasets}.
 Bold values indicate the best-performing models per metric, considering both the highest mean and overlapping standard deviations.
Across both clean and challenging datasets, CoSupFormer consistently outperforms all baselines, achieving the highest scores on nearly all evaluation metrics. On the clean TDBrain dataset (see Table \ref{tab:clean}), it delivers strong performance with 96\% accuracy and an AUROC of 99\%, significantly surpassing other transformer-based models. A similar trend is observed on the ADFTD dataset (see Table \ref{tab:clean}), where CoSupFormer slightly outperforms MedFormer on AUPRC metric while achieving comparable performance in other metrics.

Under noisy conditions, the performance gap widens, as one can see in Table \ref{tab:noisy}.
CoSupFormer maintains high robustness on the TDBrain+Noise dataset, achieving 95\% F1-score, while baseline models show substantial performance drops. On the more challenging ADFTD+Noise dataset, CoSupFormer again achieves the best results, showing clear gains over MedFormer and CrossFormer, whose performance degrades more severely. On the MACO dataset, CoSupFormer continues to outperform all baselines by a large margin, achieving over 70\% in accuracy, F1-score, and AUPRC. These results highlight the effectiveness of our approach in both standard and noisy settings, particularly in maintaining generalization to unseen subjects and corrupted signals.

\paragraph{Ablations}
To assess the contribution of key components in our architecture, we conducted ablation studies by removing the contrastive loss and the gating mechanism from CoSupFormer. These experiments help isolate the impact of each design choice on overall performance. Results of the ablations, along with detailed comparisons, are provided in the section \ref{ablation_stud} of the Supplementary Material.

\paragraph{Limitations}
This work focuses exclusively on EEG signals; while CoSupFormer shows strong performance in this domain, its applicability to other medical time series such as ECG or EMG remains untested. Additionally, in cases where the available training data is very limited, the combined supervised-contrastive loss can occasionally lead to a drop in performance, suggesting that careful tuning or loss rebalancing may be needed in low-data regimes.

\section{Conclusion}
\label{sec:c/c and limits}
We introduced CoSupFormer, a novel architecture for EEG classification that combines a multi-resolution CNN encoder, a global attention mechanism for modeling rich spatiotemporal interactions, a gating module for noise suppression, and a hybrid supervised–contrastive loss. In contrast to prior work, our model jointly captures cross-channel dynamics over time and leverages supervised and contrastive learning to achieve improved generalization. Our results demonstrate that CoSupFormer consistently outperforms existing transformer variants, achieving superior accuracy, F1-score, and AUPRC, particularly under challenging conditions where signal quality is degraded. Notably, its robustness on the noisy TDBrain, ADFTD, and MACO datasets highlights its capacity to generalize across varied signal sources, noise levels, and even species, as the datasets include both human and mouse EEG recordings. These findings suggest that CoSupFormer is a promising direction for building more reliable and context-aware EEG classifiers, especially in real-world clinical and pharmacological applications involving high subject variability and noisy recordings.

\section*{Acknowledgment}

This work was funded by SynapCell SAS through the Cortex project, awarded at the 9th edition of the i-Nov competition organized for French companies. This work was also granted access to the HPC resources of IDRIS under the allocation 2025-AD011016062 made by GENCI and the HPC ressources from GRICAD infrastructure which is supported by Grenoble research communities.

\clearpage

\bibliography{CoSupFormer_Refs}
\bibliographystyle{unsrtnat}










\newpage


\title{CoSupFormer : A Contrastive Supervised learning approach for EEG signal Classification}

\author{%
  Davy Darankoum \\
  Univ. Grenoble Alpes, CNRS, Grenoble INP, LJK, \\38000 Grenoble, France\\
  SynapCell SAS, ZAC ISIPAR, 38330 Saint-Ismier, France\\
  \texttt{davy.darankoum@univ-grenoble-alpes.fr} \\
  \And
  Chloé Habermacher \\
  SynapCell SAS, ZAC ISIPAR \\
  38330 Saint-Ismier, France \\
  \texttt{chabermacher@synapcell.fr} \\
  \And
  Julien  Volle \thanks{These authors jointly supervised the work.}\\
  SynapCell SAS, ZAC ISIPAR \\
  38330 Saint-Ismier, France \\
  \texttt{jvolle@synapcell.fr} \\
  \And
  Sergei Grudinin$^*$\\
  Univ. Grenoble Alpes, CNRS, Grenoble INP, LJK \\
   38000 Grenoble, France \\
  \texttt{sergei.grudinin@univ-grenoble-alpes.fr} \\
}



\newpage
\appendix

\counterwithin{figure}{section}
\counterwithin{table}{section}
\counterwithin{equation}{section}
\counterwithin{algorithm}{section}

\renewcommand{\thefigure}{\Alph{section}.\arabic{figure}}
\renewcommand{\theequation}{\Alph{section}.\arabic{equation}}
\renewcommand{\thealgorithm}{\Alph{section}.\arabic{algorithm}}



\section*{Appendix}



\section{Datasets Preprocessing}
Table \ref{tab:dataset_config} lists the preprocessed data characteristics according to each experiment.

\begin{table}[ht]
\centering
\scriptsize
\begin{tabular}{llcccccccc}
\toprule
\textbf{Dataset} & \textbf{Model} & \textbf{\#Subj.} & \textbf{\#Cls.} & \textbf{\#Ch.} & \textbf{\#Samples} & \textbf{Len} & \textbf{SFreq} & \textbf{BPF} & \textbf{Aug.} \\
\midrule

\multirow{3}{*}{TDBrain \& TDBrain+Noise}
    & MedFormer     & 50  & 2 & 33 & 6,240  & 256   & 256 & None & Drop0.25 \\
    & CrossFormer   & 50  & 2 & 33 & 6,240  & 256   & 256 & None & None \\
    & CoSupFormer   & 50  & 2 & 33 & 6,240  & 256   & 256 & None & None \\

\midrule

\multirow{3}{*}{ADFTD \& ADFTD+Noise}
    & MedFormer     & 88  & 3 & 19 & 69,752 & 256   & 256 & 0.5–45 & Drop0.5 \\
    & CrossFormer   & 88  & 3 & 19 & 69,752 & 256   & 256 & 0.5–45 & None \\
    & CoSupFormer   & 88  & 3 & 19 & 6,975  & 2,560 & 256 & 0.5–45 & None \\

\midrule

\multirow{3}{*}{MACO}
    & MedFormer     & 128 & 3 & 3  & 1,563,600  & 256   & 256 & None   & None \\
    & CrossFormer   & 128 & 3 & 3  & 1,563,600 & 256   & 256 & None   & None \\
    & CoSupFormer   & 128 & 3 & 3  & 26,060  & 15,360 & 256 & None  & None \\

\bottomrule
\end{tabular}
\caption{
\small 
Dataset configurations and preprocessing details.
\#Cls: number of classes. \#Ch: number of channels. Len: number of time steps per sample. SFreq: Sampling frequency in Hertz(Hz). BPF: bandpass filter in Hertz(Hz). Aug: Data augmentation. 
Augmentations applied on MedFormer experiments were introduced in \citet{wangmedformer}. Similar to the dropout layer in neural networks, this augmentation method randomly drops some values. The proportion of values dropped is controlled by the parameter \{0.25 or 0.5\}.
}
\label{tab:dataset_config}
\end{table}

\subsection{TDBrain Preprocessing}
TDBrain is a large-scale EEG time series dataset introduced by \citet{van2022two}, containing recordings from 1,274 subjects across 33 EEG channels. 
Each subject was recorded under two conditions: eyes-closed and eyes-open. The dataset includes 60 clinical labels, allowing for multi-label classification to reflect coexisting neurological conditions.

In this work, following \citet{wangmedformer}, we used a subset of the dataset consisting of 25 subjects with Parkinson’s disease and 25 healthy controls, all recorded under the eyes-closed condition. Each trial was segmented into non-overlapping 1-second samples (256 timestamps per sample), with segments shorter than one second discarded, yielding a total of 6,240 samples.
We adopted a subject-independent split: samples from subjects \{18,19,20,21,46,47,48,49\} form the validation set, those from \{22,23,24,25,50,51,52,53\} form the test set, and the remaining samples are used for training.

\paragraph{License} TDBrain dataset is made publicly available under the proprietary user data agreement that can be found at \url{https://brainclinics.com/resources/tdbrain-dataset/introduction}.

\subsection{TDBrain+Noise Preprocessing}
\label{TDBrain_Preproc}

We constructed the TDBrain+Noise dataset  by adding Gaussian noise to the last 11 of the 33 EEG channels in the original TDBrain recordings. The noise was drawn from a normal distribution with a mean of 0 and a standard deviation of 1000 $\mu$V, introducing strong signal corruption in one third of the input channels. All other aspects of preprocessing including segmentation into 1-second samples, sampling rate, and subject-independent split, were kept identical to the original TDBrain dataset.

\subsection{ADFTD Preprocessing}
\label{ADFTD_Preproc}
The Alzheimer’s Disease and Frontotemporal Dementia (ADFTD) dataset is a publicly available EEG time series collection including recordings from three subject groups: 36 patients with Alzheimer’s Disease (AD), 23 with Frontotemporal Dementia (FTD), and 29 healthy control (HC) individuals \citep{miltiadous2023dataset}. 
Each subject underwent a single EEG recording session using 19 channels, sampled at a native rate of 500 Hz. The duration of the recordings varied between groups, with average session lengths of 13.5 minutes for AD, 12 minutes for FTD, and approximately 13.8 minutes for HC. A bandpass filter between 0.5–45Hz is applied to all recordings, and signals are downsampled to 256Hz for processing.

The segmentation strategy differs across models. For experiments with MedFormer and CrossFormer, recordings are split into non-overlapping 1-second segments (256 timestamps each), and segments shorter than 1 second are discarded, yielding 69,752 samples. 
In contrast, for CoSupFormer, we use non-overlapping 10-second segments (2560 timestamps), resulting in 6,975 samples.

We used a subject-independent split:
60\% of the subjects are used for training, 20\% for validation, and 20\% for testing, ensuring no overlap between the sets.
\paragraph{License} ADFTD dataset is made publicly available under the proprietary user data agreement that can be found at \url{https://openneuro.org/datasets/ds004504/versions/1.0.6}.

\subsection{ADFTD + Noise Preprocessing}
The ADFTD+Noise dataset was created by injecting Gaussian noise into the last 6 of the 19 EEG channels from the original ADFTD recordings. The noise was sampled from a normal distribution with a mean of 0 and a standard deviation of 1000 $\mu$V, resulting in strong perturbations in the six selected channels. All other preprocessing steps including filtering, downsampling to 256Hz, segmentation, and subject-independent splitting remain identical to the original ADFTD setup.

\subsection{MACO Preprocessing}
MACO is a private dataset including 1536 hours of EEG recordings collected from 128 mice. It includes multiple pharmaco-EEG studies involving compounds from two therapeutic classes: antidepressants and antipsychotics. A third class consists of EEG recordings obtained after solvent administration, used to constitute a control condition. Recordings were obtained from three brain regions—the hippocampus, prefrontal cortex, and parietal cortex—resulting in 3-channel EEG signals.

Each recording corresponds to a 6-hour session, consisting of 1 hour before compound administration and 5 hours after. Only post-administration signals were retained for analysis. For each compound, we analyzed the power spectrum across all post-administration recordings to identify windows were EEG profiles were distinctive from the solvent. We then extracted signal segments that fell within these identified periods for all sessions.

EEG signals were acquired at a sampling rate of 1024Hz and resampled to 256Hz. For experiments using MedFormer and CrossFormer, we segmented the data into non-overlapping 1-second windows, resulting in 1,563,600 samples. For CoSupFormer, we used 60-second non-overlapping segments, yielding 26,060 samples.

We ensured that the final dataset was balanced across the three classes —- antidepressants, antipsychotics, and solvents in terms of the number of EEG samples. We split the data into training (70\%), validation (10\%), and test (20\%) sets using a subject-independent strategy to prevent data leakage across mice.

\paragraph{License} MACO dataset has been made available to us by SynapCell SAS who finance this research study.

\section{Experiments setup}
\label{Exp_setup}

All models were trained for 100 epochs using the PyTorch framework and the
Adam optimizer. 
We chose the final model according to the highest F1-score on the validation set.
For each dataset, once the subject split is determined, we run each experiment with five arbitrary seeds (41–45). 
Table \ref{tab:setup} lists further details on the model configuration for each experiment.

\paragraph{CoSupFormer} In all the experiments with CoSupFormer, we kept the dual-path encoder and global attention architectures unchanged.
Between experiments on ADFTD, ADFTD+Noise, MACO, TDBrain, and TDBrain+Noise datasets, we only varied embedding sizes. 
We also adapted the gating activation function between the experiments,
see Table \ref{tab:setup}.

\paragraph{Baseline architectures} All five baseline architectures were implemented using the raw source code made available by the respective authors. The baselines along with their associated source code linhk are : iTransformer \cite{liuitransformer} (\url{https://github.com/thuml/iTransformer}), FEDFormer \cite{zhou2022fedformer} (\url{https://github.com/MAZiqing/FEDformer}), PatchTST \cite{nietime} (\url{https://github.com/yuqinie98/PatchTST}), CrossFormer \cite{zhang2023crossformer} (\url{https://github.com/Thinklab-SJTU/Crossformer}) and MedFormer \cite{wangmedformer} (\url{https://github.com/DL4mHealth/Medformer}).

\begin{table}[ht]
\centering
\scriptsize
\begin{tabular}{llcccccccc}
\toprule
\textbf{Dataset} & \textbf{Model} & \textbf{Batch size} & \textbf{Embed-dim} & \textbf{FFN-dim} & \textbf{LR} & \textbf{Time / Params} \\
\midrule

\multirow{3}{*}{TDBrain \& TDBrain+Noise}
    & MedFormer     & 32  & 128 & 256 & 1e-4 & 35 min / 3.5M \\
    & CrossFormer   & 32  & 128 & 256 & 1e-4 & 15 min / 5.3M \\
    & CoSupFormer   & 32  & 32  & 64  & 1e-4 & 332 min / 1.07M \\

\midrule

\multirow{3}{*}{ADFTD \& ADFTD+Noise}
    & MedFormer     & 128 & 128 & 256 & 1e-4 & 180 min / 8.1M \\
    & CrossFormer   & 128 & 128 & 256 & 1e-4 & 45 min / 5.2M \\
    & CoSupFormer   & 32  & 64  & 128 & 1e-4 & 262 min / 2.4M \\

\midrule

\multirow{3}{*}{MACO}
    & MedFormer     & 20  & 128 & 256 & 1e-4 & 580 min / 3.2M \\
    & CrossFormer   & 20  & 128 & 256 & 1e-4 & 420 min / 5.2M \\
    & CoSupFormer   & 20  & 64  & 128 & 1e-3 & 271 min / 0.415M \\

\bottomrule
\end{tabular}
\caption{\small 
Training configuration and model size across datasets.
\textit{LR} denotes the learning rate. \textit{Embed-dim} refers to the embedding dimension after the dual-path encoder. \textit{FFN-dim} indicates the dimensional expansion in the feed-forward network. \textit{Time} corresponds to the average training duration in minutes across the five seeds.}
\label{tab:setup}
\end{table}


\section{Ablation studies}
\label{ablation_stud}
To better understand the contribution of each architectural component within CoSupFormer, we conducted a series of ablation studies across five datasets. These include the removal of the contrastive loss term, the gating mechanism, and the global attention module. Results are summarized in Table~\ref{tab:ablation_study}.

\subsection{Impact of CoSup loss}
To assess the contribution of the contrastive supervision loss (CoSup), we compare the full CoSupFormer model with a variant trained without the contrastive component. Across all datasets, removing the contrastive loss leads to a consistent degradation in classification performance, especially on metrics that reflect discriminative capacity such as precision, F1-score, and AUPRC.

The most pronounced difference is observed on the TDBrain+Noise dataset, where AUPRC drops by 1.57 points (from 99.34 to 97.77) and AUROC by 0.65 points. Additionally, the F1-score drops from 94.44 to 94.29, despite a small gain in precision, indicating that the model struggles more to correctly classify positive samples under noisy conditions without the additional supervision signal provided by the contrastive term.

On MACO, a dataset characterized by high variability and real-world noise, the removal of contrastive loss results in a 1-point drop in accuracy (from 78.06 to 77.06) and a 0.83-point drop in F1-score. Similar trends are observed on ADFTD, where although accuracy increases slightly (from 53.79 to 55.91), AUPRC and F1-score decline, suggesting that the gain is not consistently reflected across evaluation metrics.

These results indicate that the CoSup loss enhances the model's ability to learn robust and discriminative representations, especially under inter-subject variability and signal corruption. The auxiliary contrastive objective acts as a regularizer, promoting separation in the learned feature space between similar and dissimilar examples, which appears crucial in scenarios with limited or noisy supervision.

\subsection{Impact of Gating mechanism}

The gating mechanism proves to be an essential component of CoSupFormer, particularly in scenarios where EEG signals are noisy or contain channel-specific artifacts. On TDBrain+Noise, removing the gating module leads to a 4.7-point drop in accuracy (from 94.24 to 89.51) and a 4.5-point drop in F1-score (from 94.44 to 89.99). Likewise, on MACO, the absence of gating results in a performance decline of 4.2 points in accuracy (from 78.06 to 73.85) and 3.6 points in F1-score. These results demonstrate the gating module’s ability to suppress irrelevant or corrupted {\em patches}, helping the network focus on more informative regions and contributing to better generalization in real-world noisy conditions.

In contrast, on the ADFTD and ADFTD+Noise datasets, the performance without gating is comparable or slightly better than the full model. For example, in ADFTD, removing gating leads to +2.2 points in F1-score. This suggests that in certain settings—possibly those with more consistent spatial patterns or less inter-channel interference, the gating mechanism may be less beneficial, or even overly suppressive. These observations highlight that the effectiveness of gating is highly data-dependent and might benefit from adaptive tuning strategies in future work.

Overall, the results confirm that the gating mechanism plays a pivotal role in enhancing robustness to signal degradation, particularly in noisy or heterogeneous EEG environments such as TDBrain+Noise and MACO.

\subsection{Impact of Global Attention}
The global attention mechanism improves performance on most datasets, particularly on TDBrain and TDBrain+Noise, where its removal leads to noticeable declines in F1-score and AUPRC. These results highlight its utility in capturing long-range dependencies and cross-channel interactions in both clean and noisy EEG signals.

Interestingly, on ADFTD+Noise, the model without global attention achieves the best scores across all metrics, and on MACO, it obtains the highest AUROC and AUPRC. This suggests that in some settings, especially with highly corrupted or low-dimensional signals, the model may benefit from a simpler architecture.

Notably, even without global attention, performance remains competitive across datasets. This robustness can be attributed to the presence of the gating mechanism, which filters irrelevant features, and the contrastive loss, which encourages discriminative representations. Together, these components help maintain strong performance despite the absence of global attention.

\begin{table}[ht]
\centering
\scriptsize
\begin{tabular}{llcccccc}
\toprule
\textbf{Dataset} & \textbf{Model Variant} & \textbf{Acc.} & \textbf{Prec.} & \textbf{Recall} & \textbf{F1} & \textbf{AUROC} & \textbf{AUPRC} \\
\midrule
\multirow{4}{*}{TDBrain}
    & CoSupFormer & \textbf{96.04 ± 0.36} & \textbf{95.95 ± 1.83} & 96.18 ± 1.22 & \textbf{96.05 ± 0.30} & \textbf{99.41 ± 0.05} & \textbf{99.46 ± 0.04} \\
    & w/o Contrastive & 95.59 ± 0.22 & 94.82 ± 0.85 & \textbf{96.46 ± 0.55} & 95.63 ± 0.19 & 99.24 ± 0.06 & 99.30 ± 0.05 \\
    & w/o Gating & 90.28 ± 0.95 & 92.13 ± 1.48 & 88.13 ± 2.80 & 90.05 ± 1.12 & 96.89 ± 0.88 & 97.32 ± 0.50 \\
    & w/o Global Attention & 90.87 ± 2.05 & 89.51 ± 5.11 & 92.92 ± 1.80 & 91.09 ± 1.71 & 97.52 ± 0.37 & 97.88 ± 0.33 \\
\midrule
\multirow{4}{*}{TDBrain+Noise}
    & CoSupFormer & \textbf{94.24 ± 2.17} & 91.89 ± 4.15 & \textbf{97.22 ± 0.64} & \textbf{94.44 ± 1.93} & \textbf{99.29 ± 0.16} & \textbf{99.34 ± 0.17} \\
    & w/o Contrastive & 94.17 ± 2.00 & \textbf{93.14 ± 4.78} & 95.63 ± 1.67 & 94.29 ± 1.74 & 98.64 ± 0.53 & 97.77 ± 2.20 \\
    & w/o Gating & 89.51 ± 1.49 & 86.11 ± 1.51 & 94.24 ± 1.27 & 89.99 ± 1.40 & 97.22 ± 0.93 & 97.55 ± 0.78 \\
    & w/o Global Attention & 92.67 ± 0.96 & 90.17 ± 2.74 & 95.90 ± 1.56 & 92.91 ± 0.75 & 98.58 ± 0.16 & 98.68 ± 0.16 \\
\midrule
\multirow{4}{*}{ADFTD}
    & CoSupFormer & 53.79 ± 1.17 & 50.44 ± 0.42 & 50.71 ± 0.54 & 50.36 ± 0.31 & 70.11 ± 0.49 & 53.34 ± 0.13 \\
    & w/o Contrastive & 55.91 ± 1.72 & 52.37 ± 1.42 & 52.35 ± 0.70 & 51.74 ± 0.87 & 71.10 ± 0.51 & 52.71 ± 0.47 \\
    & w/o Gating & \textbf{60.65 ± 2.12} & \textbf{59.35 ± 4.52} & \textbf{53.33 ± 0.66} & \textbf{52.58 ± 0.70} & \textbf{71.49 ± 0.05} & \textbf{58.15 ± 0.40} \\
    & w/o Global Attention & 56.01 ± 2.87 & 55.37 ± 4.13 & 52.61 ± 2.46 & 52.49 ± 2.20 & 71.31 ± 2.13 & 57.81 ± 2.90 \\
\midrule
\multirow{4}{*}{ADFTD+Noise}
    & CoSupFormer & 52.09 ± 1.27 & 48.71 ± 1.78 & 46.90 ± 0.53 & 46.77 ± 0.61 & 66.38 ± 0.21 & 50.82 ± 0.32 \\
    & w/o Contrastive & 52.25 ± 2.17 & 50.14 ± 3.57 & 47.59 ± 0.70 & 47.34 ± 1.05 & 67.50 ± 0.60 & 51.64 ± 3.14 \\
    & w/o Gating & 52.37 ± 0.76 & 48.89 ± 1.10 & 47.79 ± 0.66 & 47.90 ± 0.73 & 66.76 ± 0.87 & 52.60 ± 0.97 \\
    & w/o Global Attention & \textbf{54.43 ± 2.25} & \textbf{59.41 ± 8.73} & \textbf{49.79 ± 1.59} & \textbf{49.62 ± 2.41} & \textbf{68.99 ± 1.36} & \textbf{55.43 ± 0.76} \\
\midrule
\multirow{4}{*}{MACO}
    & CoSupFormer & \textbf{78.06 ± 0.51} & 77.90 ± 1.59 & \textbf{76.01 ± 0.47} & \textbf{76.17 ± 0.95} & 90.28 ± 1.67 & 84.89 ± 1.37 \\
    & w/o Contrastive & 77.06 ± 3.69 & 76.86 ± 3.49 & 75.59 ± 4.15 & 75.34 ± 4.42 & 89.48 ± 2.94 & 83.23 ± 4.59 \\
    & w/o Gating & 73.85 ± 2.80 & 72.89 ± 2.05 & 74.58 ± 0.06 & 72.59 ± 1.80 & 90.76 ± 0.71 & 84.19 ± 1.49 \\
    & w/o Global Attention & 77.60 ± 0.70 & \textbf{78.13 ± 0.94} & 75.63 ± 1.00 & 75.35 ± 1.02 & \textbf{92.14 ± 1.43} & \textbf{86.66 ± 1.31} \\
\bottomrule
\end{tabular}
\caption{Ablation study of CoSupFormer variants across five datasets. Results are reported as mean ± std(standard deviation) over five seeds.}
\label{tab:ablation_study}
\end{table}

\label{app:stuff}

\section{Model Interpretation : Embeddings projection with/without Contrastive loss}
To further assess the impact of the contrastive and supervised (CoSup) loss, we visualized the latent embeddings produced by a model trained without the contrastive loss component (referred to here as SupFormer for simplicity), and by CoSupFormer, the full model incorporating the contrastive loss, using Principal Component Analysis (PCA). The projections are computed on the output embeddings immediately before the softmax activation, allowing us to examine how the models internally organize data from different classes.

\begin{figure}
    \centering
    \includegraphics[width=1\linewidth]{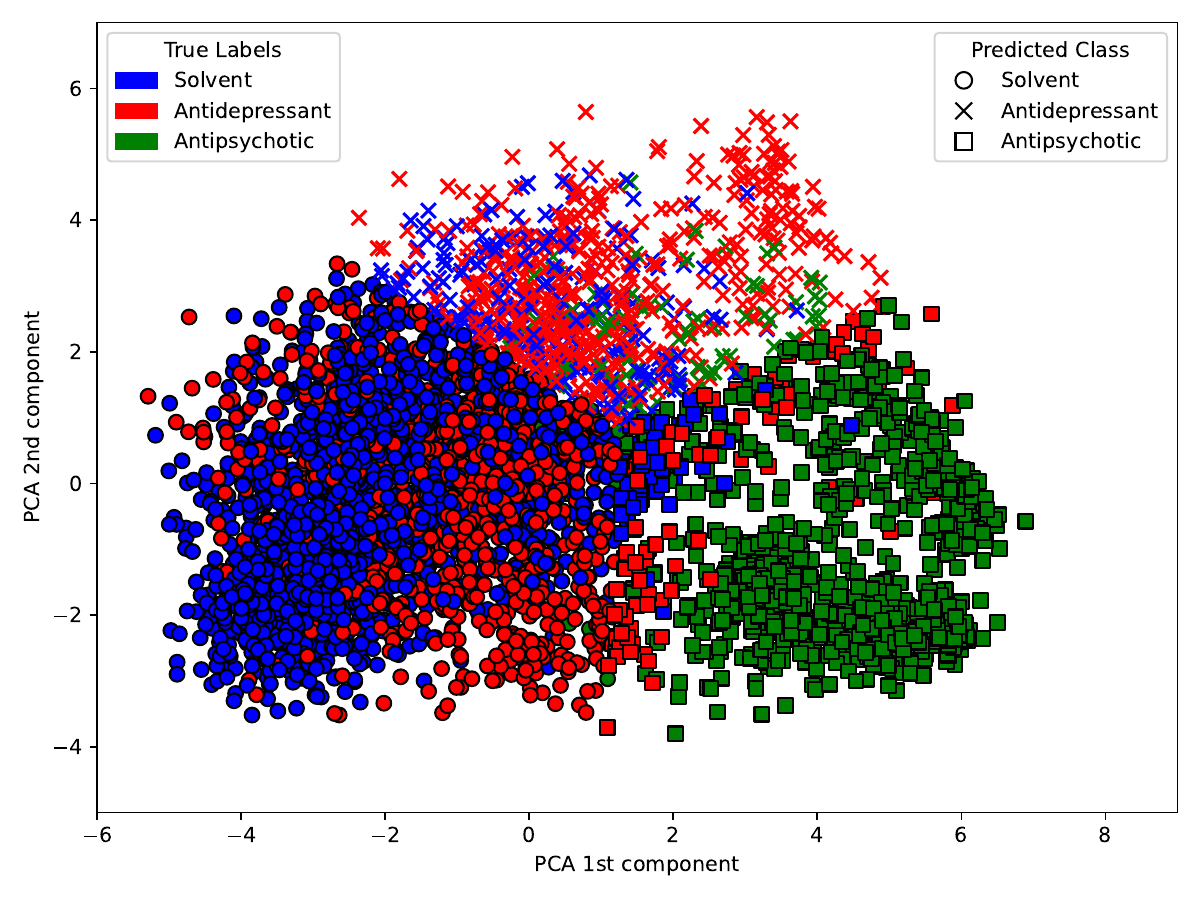}
    \caption{PCA projection of latent embeddings from the baseline model trained without contrastive loss on MACO dataset. Marker shape denotes the predicted class; color indicates the true label. Embeddings show higher intra-class variability and overlap between therapeutic categories.
    }
    \label{fig:SupFormer_42}
\end{figure}

\begin{figure}
    \centering
    \includegraphics[width=1\linewidth]{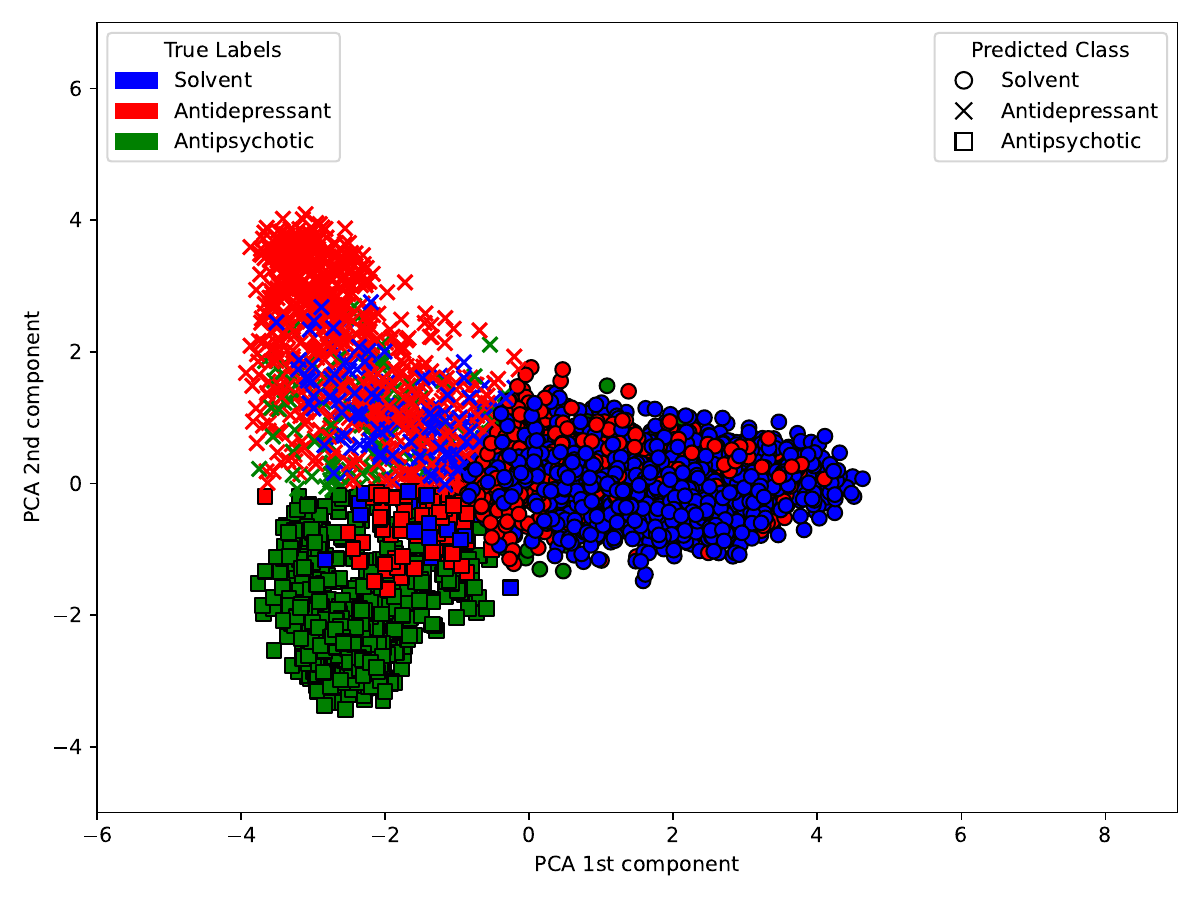}
    \caption{PCA projection of latent embeddings from CoSupFormer, trained with the contrastive and supervised (CoSup) loss on the MACO dataset. The embeddings exhibit improved class separability and alignment with ground truth labels, forming more compact clusters.
    }
    \label{fig:CoSupFormer_42}
\end{figure}

Figure~\ref{fig:SupFormer_42} and Figure~\ref{fig:CoSupFormer_42} display the PCA projections for SupFormer and CoSupFormer, respectively. In both plots, marker shape indicates the predicted class, while color encodes the true label.

The visualization reveals a marked improvement in class separability and alignment between predictions and ground truth in CoSupFormer. Specifically, the embeddings from CoSupFormer form more compact, class-specific clusters with clearer boundaries between therapeutic categories. In contrast, the SupFormer embeddings are more dispersed, with significant overlap between classes—particularly between antidepressants and antipsychotics—highlighting a reduced discriminative capacity.

These results support the quantitative findings in Table~\ref{tab:ablation_study}, demonstrating that the CoSup loss fosters a more structured and separable latent space.


\section{Broader impact}
This work contributes to the development of more robust and interpretable deep-learning models for EEG signal classification, with direct implications for clinical neuroscience and pharmacological research. By combining global attention, gating mechanisms, and a contrastive supervision, CoSupFormer is able to handle noisy, heterogeneous, and low-SNR EEG data more effectively than standard architectures.
%
CoSupFormer is particularly relevant for real-world healthcare applications where EEG recordings often suffer from artifacts and imbalanced classes. This includes contexts such as early diagnosis of neurodegenerative diseases, monitoring of psychiatric conditions, or drug efficacy evaluation in preclinical studies using animal models. Moreover, the ability to generalize across both human and animal EEG datasets supports its applicability in translational neuroscience studies.

From a societal perspective, models like CoSupFormer could eventually be integrated into clinical decision-support tools, contributing to more accessible and scalable neuro-diagnostic systems. However, the deployment of such models must be approached cautiously, with attention to validation across diverse populations, transparency of model decisions, and the ethical implications of automated predictions in sensitive medical settings.
%
Overall, this research aligns with the broader goal of making AI-driven neuro-technology more robust, equitable, and clinically useful.

\clearpage




\end{document}